# CLASS ATTENDANCE SYSTEM IN EDUCATION WITH DEEP LEARNING METHOD

Hüdaverdi DEMİR[1,2], Serkan SAVAŞ[3]

[1] *Department of Electronics and Computer Engineering, Graduate School of Natural and Applied Sciences, Çankırı Karatekin University, Çankırı, Türkiye*
[2] *Department of Informatics Technology, Yenikent Ahmet Çiçek Vocational and Technical Anatolian High School, Ankara, Türkiye*
[3] *Department of Computer Engineering, Faculty of Engineering and Architecture, Kırıkkale University, Kırıkkale, Türkiye*

## Abstract

With the advancing technology, the hardware gain of computers and the increase in the processing capacity of processors have facilitated the processing of instantaneous and real-time images. Face recognition processes are also studies in the field of image processing. Facial recognition processes are frequently used in security applications and commercial applications. Especially in the last 20 years, the high performances of artificial intelligence (AI) studies have contributed to the spread of these studies in many different fields. Education is one of them. The potential and advantages of using AI in education; can be grouped under three headings: student, teacher, and institution. One of the institutional studies may be the security of educational environments and the contribution of automation to education and training processes. From this point of view, deep learning methods, one of the sub-branches of AI, were used in this study. For object detection from images, a pioneering study has been designed and successfully implemented to keep records of students' entrance to the educational institution and to perform class attendance with images taken from the camera using image processing algorithms. The application of the study to real-life problems will be carried out in a school determined in the 2022-2023 academic year.

## Introduction

With the advancing technology, the hardware gain of computers and the increase in the processing capacity of processors have facilitated the processing of instantaneous and real-time images. In this way, studies in the field of image processing have increased rapidly. The images taken from the cameras are processed on hardware-powered computers and researchers have done many studies on topics such as (Uçar, 2019):

- Recognition of the visible object,
- Face detection and recognition,
- Tracking the face,
- Recognizing human emotions,

---

[1] MSc. Candidate, Çankırı Karatekin University, Department of Electronics and Computer Engineering,
*E-mail address: hudaverdidemir@gmail.com*
[2] Informatics Technology Teacher, Yenikent Ahmet Çiçek Vocational and Technical Anatolian High School.
[3] Assist. Prof. Dr., Kırıkkale University, Department of Computer Engineering,
*E-mail address: serkansavas@kku.edu.tr*

• Human gender and age determination,

• Sign language detection.

The application of the face recognition technique is divided into two main parts security applications and commercial applications. Facial recognition technology has primarily been used in security applications, especially in photo albums for criminal records, and in video surveillance (real-time matching with video footage sequences). Commercial applications range from the static matching of photos on credit cards, ATM cards, passports, driver's licenses, and photo IDs to real-time matching with still images or video image sequences for access control. Each application has different restrictions in terms of processing information and obtaining results (Tolba et al., 2006).

In educational institutions, systems are designed with student safety in mind. Many technological methods such as Radio Frequency Identification (RFID), wireless communication, fingerprint, iris, and advanced face recognition based, etc. are tested and developed in security systems. Most of these methods have high system installation costs and have some advantages and disadvantages. Considering that school budgets are not too high, an automatic attendance system was designed using the existing technological infrastructure without the cost of extra equipment.

Especially in the last 20 years, the high performances of artificial intelligence (AI) studies have contributed to the spread of these studies in many different fields. Education is one of them. The potentials and advantages of using AI in education can be grouped under three headings such as students, teachers, and institutions (Savaş, 2021). One of the institutional studies may be the security of educational environments and the contribution of automation to education and training processes. From this point of view, deep learning (DL) methods, one of the sub-branches of AI, were used in this study. In the study, image processing algorithms were used for object detection from images, student admission records were kept with the images taken from the camera, and a pioneering study was designed to perform class attendance.

## Literature Review

Eldem and Palalı (2017) used Open Source Computer Vision Library (OpenCV) and image processing libraries in their study. In the study, the OpenCVSharp component, which was developed for the C # programming language that works in harmony with OpenCV, was used. In the system, images of the people were taken using the camera and the facial regions were marked with the haarcascade structure. The faces registered in the database and the faces from the camera were compared using the template matching method. In this study, a success rate of 79% was achieved in face recognition (Eldem, Eldem, & Palalı, 2017).

Kaplan (2018) used the Haar-Cascades classifier to determine whether there is a face in any image. In the study, AForge.NET software library was used to speed up image processing. EmguCV software library, which uses timer logic, was used to prevent lag in the interface. As a result, in this thesis study, face detection in images has been successfully performed in real time to meet the needs of the system (Kaplan, 2018).

In his thesis, Uçar (2019) determined the distraction rates of students in a classroom environment where their faces are recognized in real time, their head directions are followed and their head direction movements are interpreted. He used OpenCV and Dlib image processing and machine learning libraries in his system. In the developed application, different head directions and facial expressions of the students were photographed and recorded. Using the Local Binary Patterns method of this training dataset, the students' face recognition model was created. Photos labeled as "Careful" and "Careless" were detected using the support vector machine algorithm. As a result of the tests, the success rate of the system was determined as 72.4% (Uçar, 2019).

In the study by Savaş et al. (2017), it was aimed to find how many human faces are in the photographs taken by a mobile device and to calculate the occupancy capacity of the environment with this number. In the study, haarcascade_frontalface_alt and haarcascade_mcs_eyepair_big algorithms were used to detect face and eyes in photographs taken with smart devices. As a result, the haarcascade_frontalface_alt algorithm works more efficiently in the study (Savaş, Becerikli, & İlkin, 2017).

## Method

The architecture of the study is shown in Figure 1.

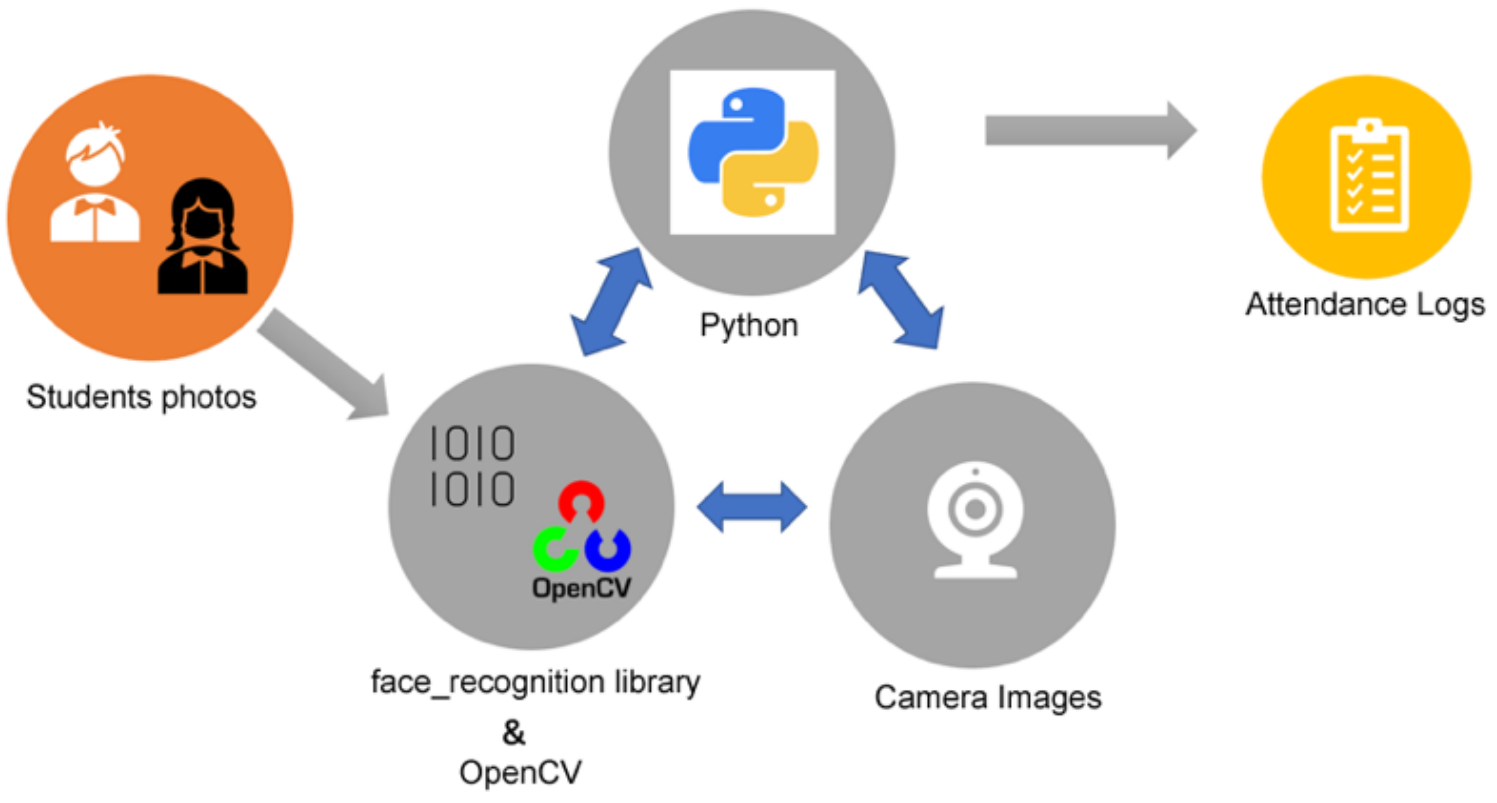

**Figure 1.** Architecture of the study

As seen in Figure 1, the images of the students are taken with the camera at the entrance of the classroom, and the face of the student is detected by using image processing algorithms, one of the DL methods. The image data obtained is compared with the class database and the student's attendance information is recorded. The steps for this process are as follows:

- Creating dataset from student images (by class)
- Obtaining lecture attendance image via the camera
- Face detection using Convolutional Neural Network (CNN) and Histogram of Oriented Gradients (HOG)
- Face identification
- Registration of attendance records as a result of face recognition

*Dataset*

Before applying the designed study to real-world problems, design and development studies were carried out. A dataset consisting of web photos of real users and different popular names has been created, which is easy to access during the development phase and provides the opportunity to act according to the situation. In the study, the comparison of this data set, which was created with the image to be taken from the camera, was carried out.

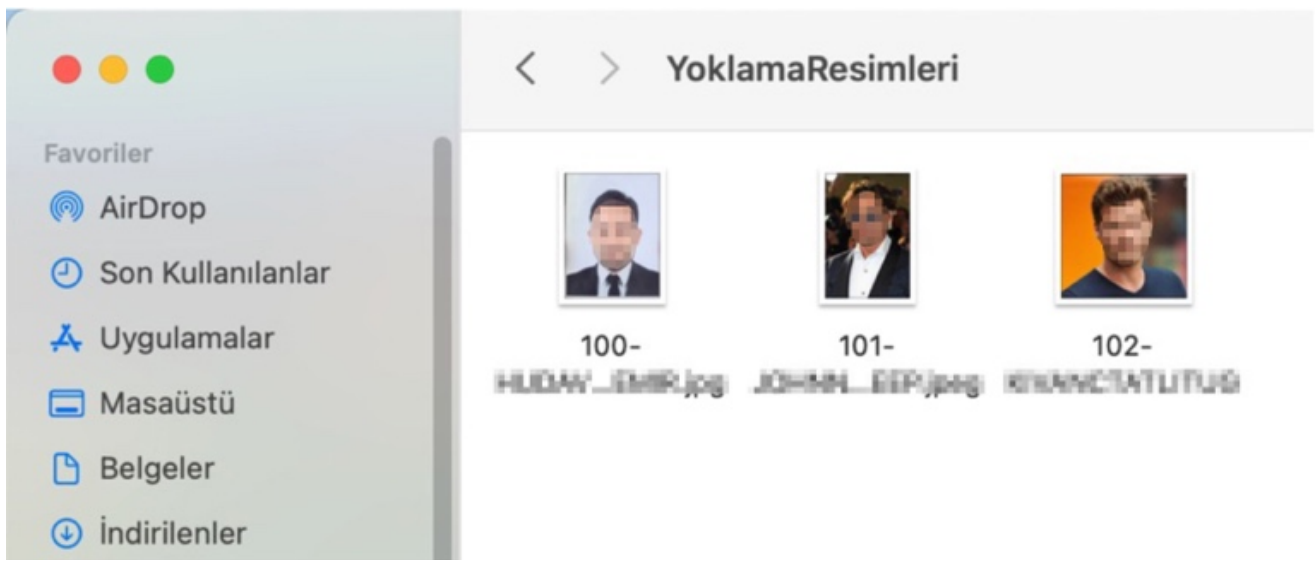

**Figure 2.** Sample images of the dataset

The images added to the data set were recorded as id and the person's name and surname. To prevent more than one person from entering the camera angle in the study, the resolution is set to 150x150 pixels at the beginning.

*Face Detection Technique*

Face detection is used to define this information by finding the coordinates and size of face objects in an image or video sent to the system.

In face detection processes, some problems are usually encountered in images obtained from uncontrolled environments. These problems can be listed as (Marques, 2010):

- Exposure variation: The ideal scenario for face detection is one that only includes frontal images, but in general uncontrolled conditions this is unlikely. Also, the performance of face detection algorithms degrades drastically when there are large pose variations. There may also be a change in pose due to the person's movements or the angle of the camera.
- Attribute occlusion: The presence of elements such as a beard, glasses, or hat provides high variability. Faces can also be partially covered by objects or other faces.
- Facial expression: Facial features also vary greatly due to different facial movements.
- Viewing conditions: Different cameras and environmental conditions can affect the appearance of a face, which can affect the quality of the image.

In order to increase the accuracy of the face recognition system, the accuracy rate of the face detection algorithm to be used should also be the highest. If faces are not detected correctly in the system, problems may occur during polling, malfunction, or the system may have to be restarted.

In the system we have designed for educational institutions to use in real-world problems, these problems will be taken under control in general, since the student photos to be included in the dataset will be taken by the institution. Apart from the problems that were overcome to create the dataset, students will also be warned so that the camera can get a good view during the polling process.

The basic information processing process in the real-time polling system was created using the Python programming language and utilizing the face-recognition library. This library is built using the face recognition feature of the Dlib library and is built with DL. The accuracy of the model reached 99.38% in different studies (Geitgey, 2020). Dlib includes machine learning (ML) and DL algorithms (Recursive Least Squares, Support Vector Machine, K-Means, CNN, Deep Neural Network, Artificial Neural Network, Sequential Minimal Optimization) and tools (Speeded Up Robust Features, HOG, Fast HOG, Color Space Conversions) to build complex software in C++ to solve real-world problems. It is an open-source library. It is used in both industry and academia in a wide variety of fields, including robotics, embedded devices, mobile phones, and large high-performance computing environments (Dlib, 2022; Pişkin, 2018).

Open Source Computer Vision Library (OpenCV), which will be used in the designed architecture, is an open source computer vision and ML software library. The library has more than 2500 optimized algorithms that include a comprehensive set of computer vision and ML algorithms. These algorithms can be used to detect and recognize faces, identify objects, classify human actions in videos, monitor camera movements, track moving objects, extract 3D models of objects, generate 3D point clouds from stereo cameras, and combine images to achieve high resolution (OpenCV, 2022).

There are two different face detection models in the Face-recognition library to be used in the designed architecture. These are HOG and CNN algorithm used for DL-based face detection. In this study, face detection was performed using HOG. CNN, on the other hand, is widely used in video processing applications where there are many video frames to be processed. CNN can be up to 3x faster for batch processing if large numbers of images are to be processed and a graphics processing unit (GPU) with Compute Unified Device Architecture (CUDA) is used. However, the HOG model, which is faster for both hardware costs and less processing in educational institutions, was preferred in this study.

*Face Detection with HOG*

Face detection is one of the most challenging problems in ML. The HOG used in the study is a feature descriptor used in machine vision for processing digital images to detect objects. HOG is also widely used in detecting moving objects (Aditya et al., 2022). Feature extraction using HOG and Gradient Direction and Gradient Magnitude is shown in Figure 3.

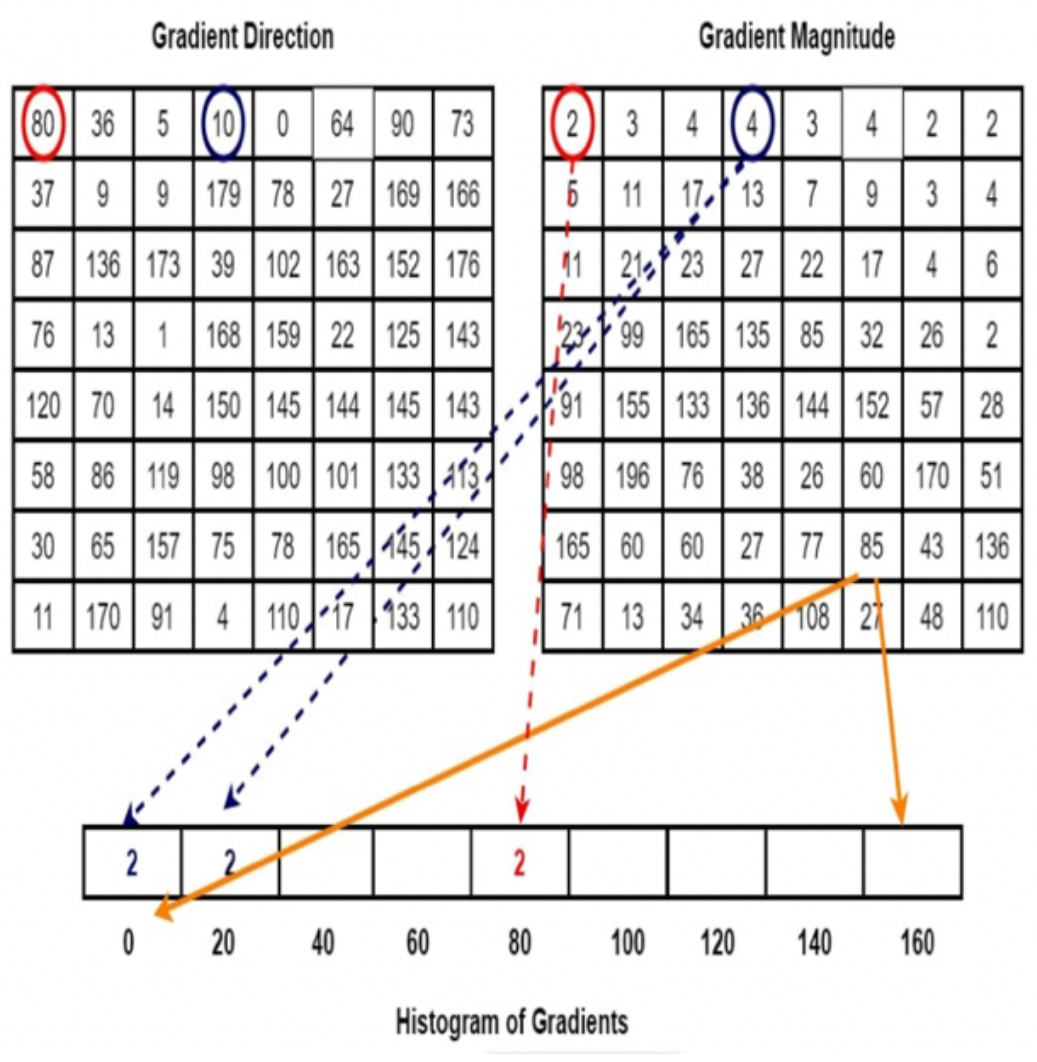

**Figure 3.** Feature extraction using HOG (Aditya et al., 2022)

Direction and magnitude calculation formulas are given in Equation (1) and Equation (2), respectively (Aditya et al., 2022).

$$Direction: \theta = tan^{-1}\left(\frac{\partial f}{\partial y} / \frac{\partial f}{\partial x}\right) \tag{1}$$

$$Magnitude: ||\nabla f|| = \sqrt{\left(\frac{\partial f}{\partial x}\right)^2 + \left(\frac{\partial f}{\partial y}\right)^2} \tag{2}$$

After successfully calculating the gradient direction and magnitude for each pixel using Equation (1) and Equation (2), features are extracted as shown in Figure 4 for example.

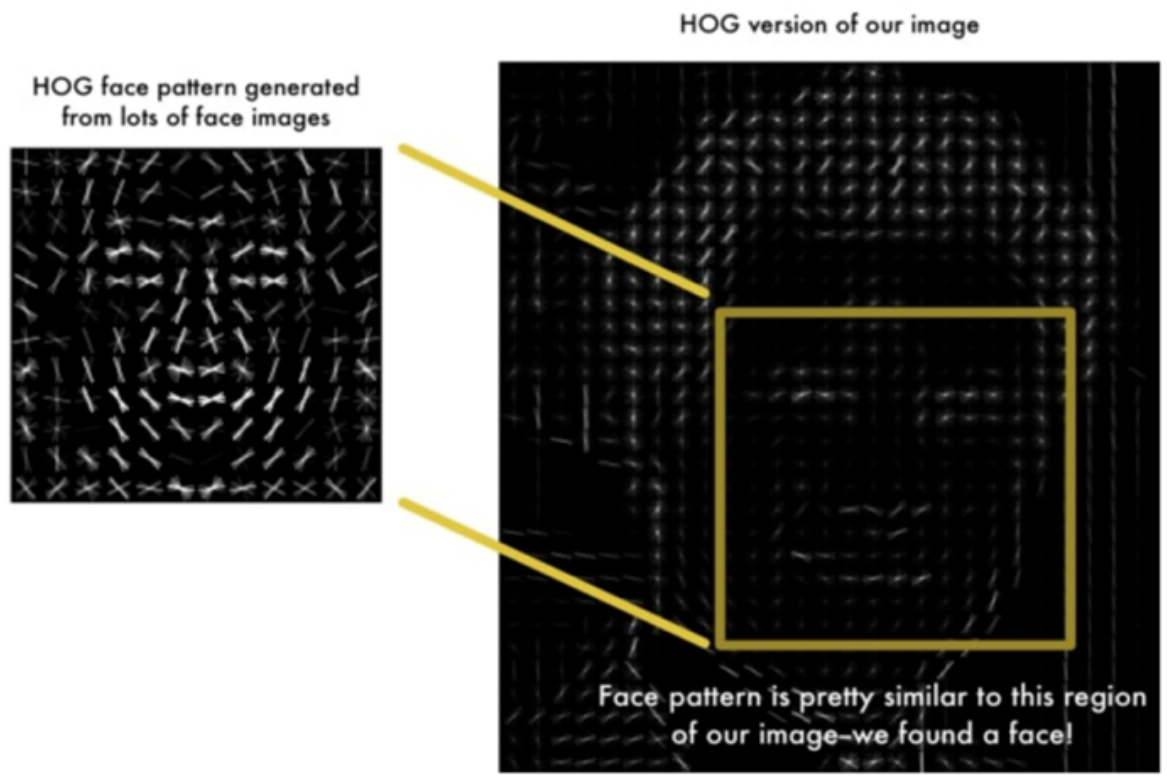

**Figure 4.** Example of image detection using HOG (Aditya et al., 2022)

*Face Recognition*

In the face recognition phase, the important measurements of the face area of all the photographs in the dataset were recorded with the face recognition algorithm. These measurements are a set of Red/Green/Blue (RGB) values for the algorithm learned only from the data samples provided to it. The algorithm used for face recognition notes some important metrics on the face such as the color, size, and slope of the eyes, the gap between the eyebrows, etc. All this together defines the face coding (information from the image) used to identify the particular face. Face coding consists of 128 numbers. Each of these numbers represents an orthogonal component of face encoding. Figure 5 shows sample values obtained from faces in the dataset.

```
array([-0.10213576,  0.05088161, -0.03425048, -0.09622347, -0.12966095,
        0.04867411, -0.00511892, -0.03418527,  0.2254715 , -0.07892745,
        0.21497472, -0.0245543 , -0.2127848 , -0.08542262, -0.00298059,
        0.13224372, -0.21870363, -0.09271716, -0.03727289, -0.1250658 ,
        0.09436664,  0.03037129, -0.02634972,  0.02594662, -0.1627259 ,
       -0.29416466, -0.12254384, -0.15237436,  0.14907973, -0.09940194,
        0.02000656,  0.04662619, -0.1266906 , -0.11484023,  0.04613583,
        0.1228286 , -0.03202137, -0.0715076 ,  0.18478717, -0.01387333,
       -0.11409076,  0.07516225,  0.08549548,  0.31538364,  0.1297821 ,
        0.04055009,  0.0346106 , -0.04874525,  0.17533901, -0.22634712,
        0.14879328,  0.09331974,  0.17943285,  0.02707857,  0.22914577,
       -0.20668915,  0.03964197,  0.17524502, -0.20210043,  0.07155308,
        0.04467429,  0.02973968,  0.00257265, -0.00049853,  0.18866715,
        0.08767469, -0.06483966, -0.13107982,  0.21610288, -0.04506358,
       -0.02243116,  0.05963502, -0.14988004, -0.11296406, -0.30011353,
        0.07316103,  0.38660526,  0.07268623, -0.14636359,  0.08436179,
        0.01005938, -0.00661338,  0.09306039,  0.03271955, -0.11528577,
       -0.0524189 , -0.11697718,  0.07356471,  0.10350288, -0.03610475,
        0.00390615,  0.17884226,  0.04291092, -0.02914601,  0.06112404,
        0.05315027, -0.14561613, -0.01887275, -0.13125736, -0.0362937 ,
        0.16490118, -0.09027836, -0.00981111,  0.1363602 , -0.23134531,
        0.0788044 , -0.00604869, -0.05569676, -0.07010217, -0.0408107 ,
       -0.10358225,  0.08519378,  0.16833456, -0.30366772,  0.17561394,
        0.14421709, -0.05016343,  0.13464174,  0.0646335 , -0.0262765 ,
        0.02722404, -0.06028951, -0.19448066, -0.07304715,  0.0204969 ,
       -0.03045784, -0.02818791,  0.06679841])
```

**Figure 5.** Sample face coding (Solegaonkar, 2019)

In the study, the similarities between the faces were determined in the next step. Each component of all compared faces is checked and it is checked whether the component at hand changes within the tolerance limits. The two sequences seen in Figure 6 indicate the similarity of the given image (in the second parameter) with each of the known face encodings in the provided list (in the first parameter). It is seen that the first sequence in the figure shows much more similarity, indicating that it accurately describes the person (Solegaonkar, 2019).

```
[array([ True,  True,  True,  True,  True,  True,  True,  True,  True,
         True,  True,  True,  True,  True,  True,  True,  True,  True,
         True,  True,  True,  True,  True,  True,  True,  True,  True,
         True,  True,  True,  True,  True,  True,  True,  True,  True,
         True,  True,  True,  True, False,  True,  True,  True,  True,
         True,  True,  True,  True,  True,  True,  True,  True, False,
         True,  True,  True,  True,  True,  True,  True,  True,  True,
         True,  True,  True,  True,  True, False,  True,  True,  True,
         True,  True,  True,  True,  True,  True,  True,  True, False,
         True,  True,  True,  True,  True,  True,  True,  True,  True,
         True,  True,  True, False,  True,  True,  True,  True,  True,
         True,  True,  True,  True,  True,  True,  True,  True,  True,
         True,  True,  True,  True,  True,  True,  True,  True,  True,
         True,  True,  True,  True,  True,  True,  True,  True, False,
         True,  True]),
 array([ True,  True,  True,  True,  True,  True, False, False, False,
         True,  True,  True, False,  True,  True,  True, False,  True,
        False,  True,  True,  True,  True, False,  True,  True,  True,
        False,  True,  True,  True, False,  True,  True,  True,  True,
         True,  True,  True,  True, False,  True, False,  True,  True,
         True,  True,  True, False,  True, False,  True,  True,  True,
        False, False,  True,  True,  True,  True,  True, False,  True,
        False, False, False, False,  True, False,  True, False,  True,
        False,  True,  True,  True,  True, False,  True,  True,  True,
         True,  True,  True, False,  True,  True,  True, False,  True,
         True, False,  True,  True,  True,  True,  True,  True,  True,
         True,  True,  True,  True,  True, False, False,  True,  True,
        False, False, False,  True,  True, False,  True,  True,  True,
         True,  True,  True,  True,  True,  True, False, False,  True,
         True,  True])]
```

**Figure 6.** Code similarities of faces (Solegaonkar, 2019)

If multiple matches are provided for the same person, people in the dataset may look very similar to each other in the photos. In such a case, a lower tolerance value is required to make face comparisons more stringent. More accurate identifications can be made by changing the value of the tolerance parameter. The default tolerance value is 0.6. Lower numbers make face comparisons more stringent. As a result of the comparisons, the login records are stored in a .csv file for later use (Face Recognition, 2017).

## 1. Results and Discussion

The performances of two different face detection models (HOG and CNN) in the Face-recognition library used in the study were also compared. In the experiments carried out, it was concluded that the HOG model captures the image on the camera faster and the camera frame per second (FPS) speed works better. Image capture rate and FPS rates are shown in Table 1.

**Table 1.** Image capture performance results

| Model Name | Image Capture Speed (Sec) | Fps Value |
|---|---|---|
| HOG | 0.0322 | 15 |
| CNN | 0.3262 | 5 |

When the image was obtained with the system designed in the study, all the photos in the data set were encoded and the facial features were kept in a list. Then, when the camera compares the detected face with the features of the faces in the data set and provides a match, the person's id and name-surname information is shown on the screen. If the match is not achieved, the information that the face is not recognized is displayed on the screen. Examples of identified (matched) and unidentified (unmatched) faces are shown in Figure 7, respectively.

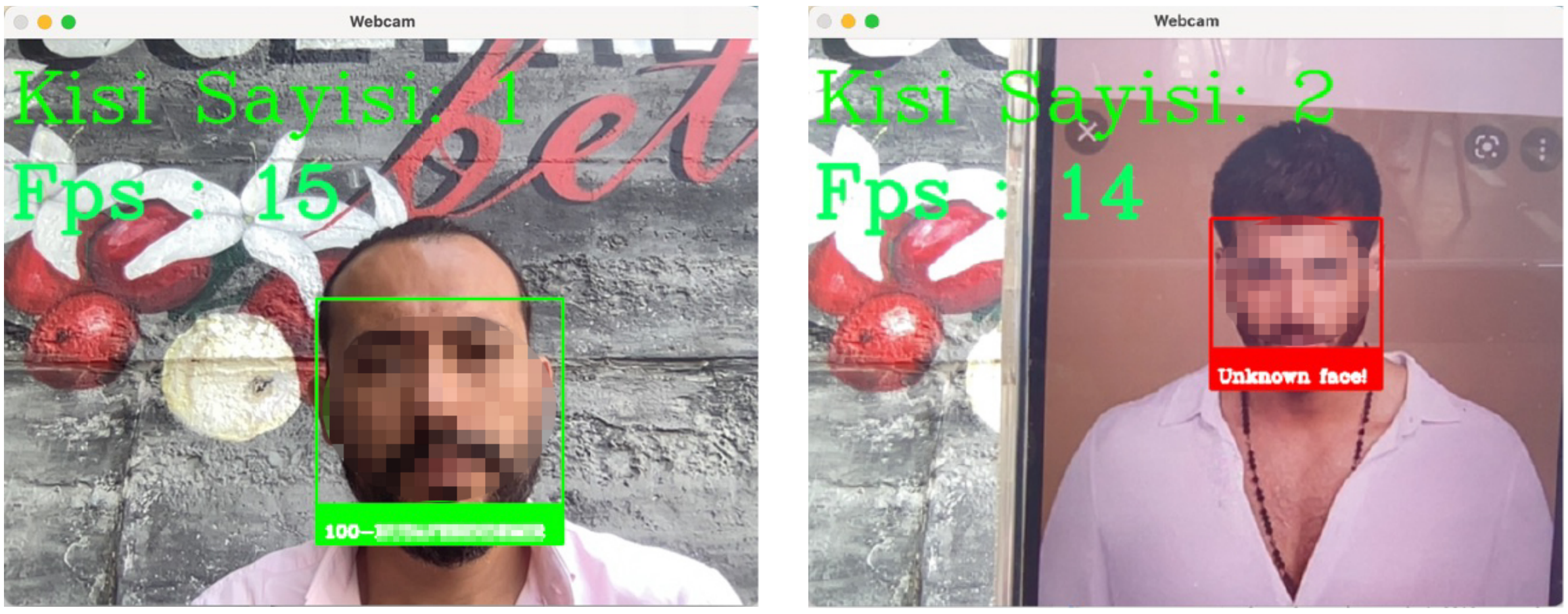

(a) (b)

**Figure 7.** (a) Identified and (b) unidentified face image

In the study, student attendance information is recorded in a .csv file. The information obtained was compared with the list previously entered into the database, and attendance-absence information was kept. As seen in Figure 8 in the sample attendance list, the id and name-surname, time, date and status information of the individuals can be checked by the teacher.

output

| AdSoyad | Zaman | Tarih | Durum |
|---|---|---|---|
| 102-[redacted] | | 2022-07-13 | yok |
| 100-[redacted] | 15:53:55 | 2022-07-13 | var |
| 101-[redacted] | 16:55:23 | 2022-07-13 | var |

**Figure 8.** Student attendance results

Although this designed system is not a new study for face recognition studies, it sets an example in terms of its adaptability to class attendance processes in education and training environments. In addition to class attendance, it is also important for security purposes in field-based education institutions such as Vocational and Technical Anatolian High Schools, when students enter and exit the workshops and laboratories in their own fields.

Automating time-consuming attendance, especially in crowded classrooms, will be an important innovation for teachers. In addition, it will also contribute to the more efficient use of FATIH hardware infrastructure established in all schools in Turkey.

## 2. Conclusions

AI technologies, which have started to be used in many different disciplines, have also started to be used in educational environments today. Equipping educational environments with automated systems is one of these uses. In this study, face recognition and polling system have been developed in educational institutions during the testing phase to set an example for and pave the way for automatic class attendance systems. The system established in the study successfully detected the faces in the data set and successfully recorded the absenteeism of the names in the class list into the database. This study helped to understand and apply the presented face recognition system to take it one step further using a real-time application.

The application of the study to real-life problems will be carried out in a school determined in the 2022-2023 academic year. With the results obtained and the feedback received from this conference, an article study will also be carried out and the results will be disseminated. In addition, steps will be taken toward the use of innovative technologies in education.

---

**HÜDAVERDİ DEMİR BIODATA**

He is MSc student in Çankırı Karatekin University, Electronics and Computer Engineering Department. He is also a teacher at Yenikent Ahmet Çiçek Voc. and Tech. Anatolian High School. Previously, he worked as a teacher and administrator in schools affiliated to the TR Ministry of National Education. He took part in national and international projects in partnership with institutions such as TUBITAK and the National Agency. He teaches Robotics and Coding, Object-Oriented Programming, Web Design courses in his institution.

**SERKAN SAVAŞ BIODATA**

He is a faculty member at Kırıkkale University. He has written and managed many national and international projects such as artificial intelligence and deep learning applications in education, 3D technologies, entrepreneurship, social and local purpose projects. His has studies and gives courses on artificial intelligence, machine learning, deep learning, cyber security, data mining, big data, software engineering, social media analysis, etc.